\documentclass[10pt,twocolumn,letterpaper]{article}
\pdfoutput=1
\usepackage{iccv}
\usepackage{times}
\usepackage{epsfig}
\usepackage{graphicx}
\usepackage{amsmath}
\usepackage{amssymb}
\usepackage{multirow}

\usepackage[breaklinks=true,bookmarks=false]{hyperref}
\usepackage{hyperref}
\hypersetup{
    colorlinks=true,
    linkcolor=blue,
    filecolor=magenta,      
    urlcolor=magenta,
}

\iccvfinalcopy 

\def\iccvPaperID{****} 

\ificcvfinal\fi

\begin{document}

\title{Visually Guided Sound Source Separation and Localization\\ using Self-Supervised Motion Representations}

\author{Lingyu Zhu\\
Computer Vision Group\\
Tampere University, Finland\\
{\tt\small lingyu.zhu@tuni.fi}
\and
Esa Rahtu\\
Computer Vision Group\\
Tampere University, Finland\\
{\tt\small esa.rahtu@tuni.fi}
}

\maketitle
\ificcvfinal\thispagestyle{empty}\fi

\begin{abstract}
   The objective of this paper is to perform audio-visual sound source separation, i.e.~to separate component audios from a mixture based on the videos of sound sources. Moreover, we aim to pinpoint the source location in the input video sequence. Recent works have shown impressive audio-visual separation results when using prior knowledge of the source type (e.g. human playing instrument) and pre-trained motion detectors (e.g. keypoints or optical flows). However, at the same time, the models are limited to a certain application domain. In this paper, we address these limitations and make the following contributions: i) we propose a two-stage architecture, called Appearance and Motion network (AMnet), where the stages specialise to appearance and motion cues, respectively. The entire system is trained in a self-supervised manner; ii) we introduce an Audio-Motion Embedding (AME) framework to explicitly represent the motions that related to sound; iii) we propose an audio-motion transformer architecture for audio and motion feature fusion; iv) we demonstrate state-of-the-art performance on two challenging datasets (MUSIC-21 and AVE) despite the fact that we do not use any pre-trained keypoint detectors or optical flow estimators. Project page: \href{https://ly-zhu.github.io/self-supervised-motion-representations}{https://ly-zhu.github.io/self-supervised-motion-representations}.
\end{abstract}

\section{Introduction}

Sound source separation is a classical task of extracting a target sound source from a given audio mixture \cite{ghahramani1996factorial,roweis2001one,virtanen2007monaural,cichocki2009nonnegative}. A well-known example is so called cocktail party problem, where one attempts to listen to a person while multiple people are speaking in the same space. Similarly, one might be interested in extracting the sound of a single instrument from a concert recording or other signals from the background noise. Despite being extensively studied, the audio-based source separation remains a challenging problem. 

Recent works \cite{ma2009lip,golumbic2013visual,ephrat2018looking,gao2018learning,owens2018audio,zhao2018sound,zhao2019sound,xu2019recursive,gao2019co,zhu2020visually,zhu2020separating} have shown that visual observations of the sound source (e.g. speaking face) may substantially simplify the separation task. For instance, the lip movements can be applied to extract the desired speech signal in the cocktail party problem \cite{ma2009lip,golumbic2013visual,ephrat2018looking}. Similarly, other visual cues like object categories or motions can be exploited to guide the separation process  \cite{gao2018learning,owens2018audio,zhao2018sound,zhao2019sound,xu2019recursive,gao2019co,zhu2020visually,zhu2020separating}. This kind of problem setup is often referred as visual sound source separation or visually guided sound source separation. 
\begin{figure}
    \centering
    \includegraphics[width=0.85\linewidth]{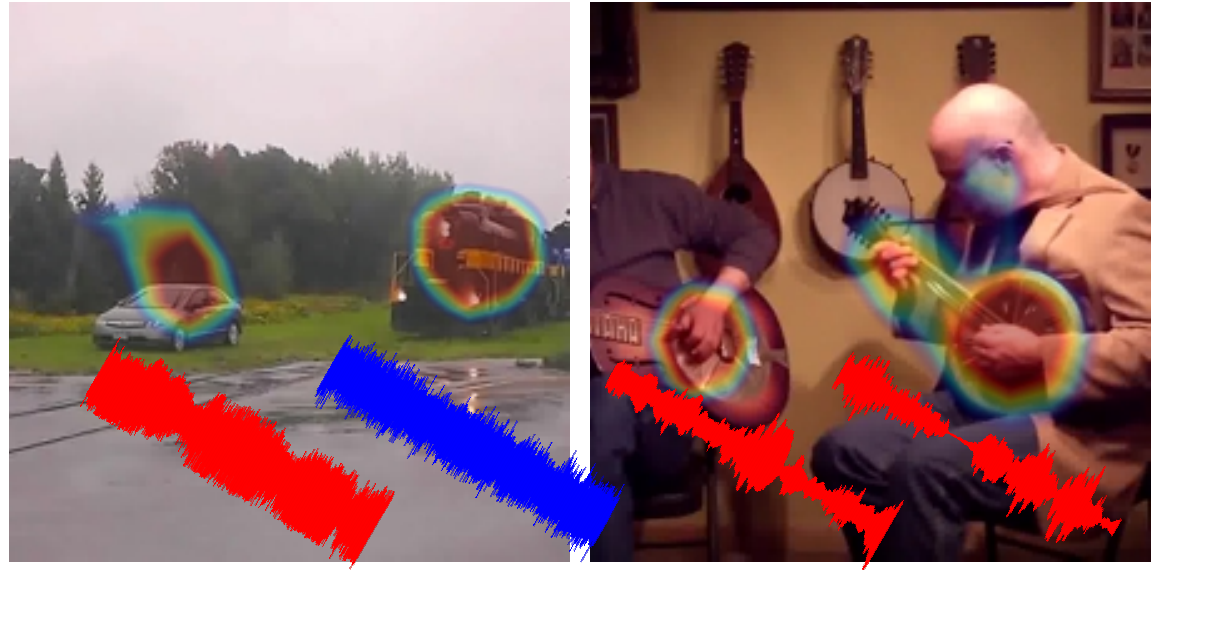}
   \caption{We introduce a new sound-related motion representation for localizing and separating sound sources. When combined with the appearance cues, we obtain a highly effective model for the visual sound separation.}
\label{fig:fig1}
\end{figure}

A popular approach in visual sound source separation is to encode both visual input and mixture audio into feature representations, and then fuse them to decode the component signal corresponding to the visual content. The visual feature encoding is one of the key elements in the approach, and previous works have studied different options for this part. For instance, several works utilise frame-based appearance features~\cite{zhao2018sound,gao2018learning,gao2019co,xu2019recursive,zhu2020separating}, which contain information about the object categories (e.g.~instrument types) but not from the object motion. While the appearance alone can be a strong cue, the motion may often be the only reliable cue (e.g. lip motion while speaking). 

Several recent works have proposed approaches to include motion information in the separation process. These models utilise optical flow~\cite{zhao2019sound}, dynamic image~\cite{zhu2020visually}, and keypoint-based human body dynamics~\cite{gan2020music}. While optical flow and dynamic images are good in representing the overall motion (e.g. for action recognition), they do not provide significant gain over the pure appearance-based methods in source separation. The recent work by Gan~\textit{et al.}~\cite{gan2020music} proposed to use human keypoints to encode the motion cues. Their main motivation was to explicitly model body and finger movements of musicians when they perform music. The results demonstrated impressive performance over the prior state-of-the-art, but unfortunately this approach is limited to sounds resulting from human motion. Moreover, the work relies on pre-trained human keypoint detectors. Owens~\textit{et al.}~\cite{owens2018audio} presented an alternative approach for learning the visual representation by classifying artificially misaligned audio and visual streams. Their idea is interesting as the misalignment naturally encourages to focus on the correlation between motion and audio.

In this paper, we introduce a two-stage visual sound source separation architecture, called Appearance and Motion network (AMnet), where the stages specialise to appearance and motion cues, respectively. We propose an Audio-Motion Embedding (AME) framework to learn the motions of sounds in a self-supervised manner. Furthermore, we design a new Audio-Motion Transformer (AMT) module to facilitate the fusion of audio and motion cues.  

We demonstrate the performance of the proposed AMnet with two challenging datasets, \textit{MUSIC-21} and \textit{AVE}, and obtain the state-of-the-art results. Interestingly, AMnet outperforms the keypoint-based approach \cite{gan2020music} without any prior knowledge or limitations to human induced motion or pre-extracted keypoints. Moreover, we apply the learned motion cues from the AME to pinpoint the sound source location in the input video stream (see Figure~\ref{fig:fig1}).

In summary, our contributions are: i) AMnet for self-supervised visual sound separation. The approach makes no pre-assumptions on the sound source type (e.g. human induced motion); ii) self-supervised motion representation learned by mapping the audio and motions into a common embedding space; iii) audio-motion transformer module for audio and motion feature fusion; iv) state-of-the-art results in two challenging visual sound separation datasets.

\begin{figure*}
    \centering
    \includegraphics[width=0.90\linewidth]{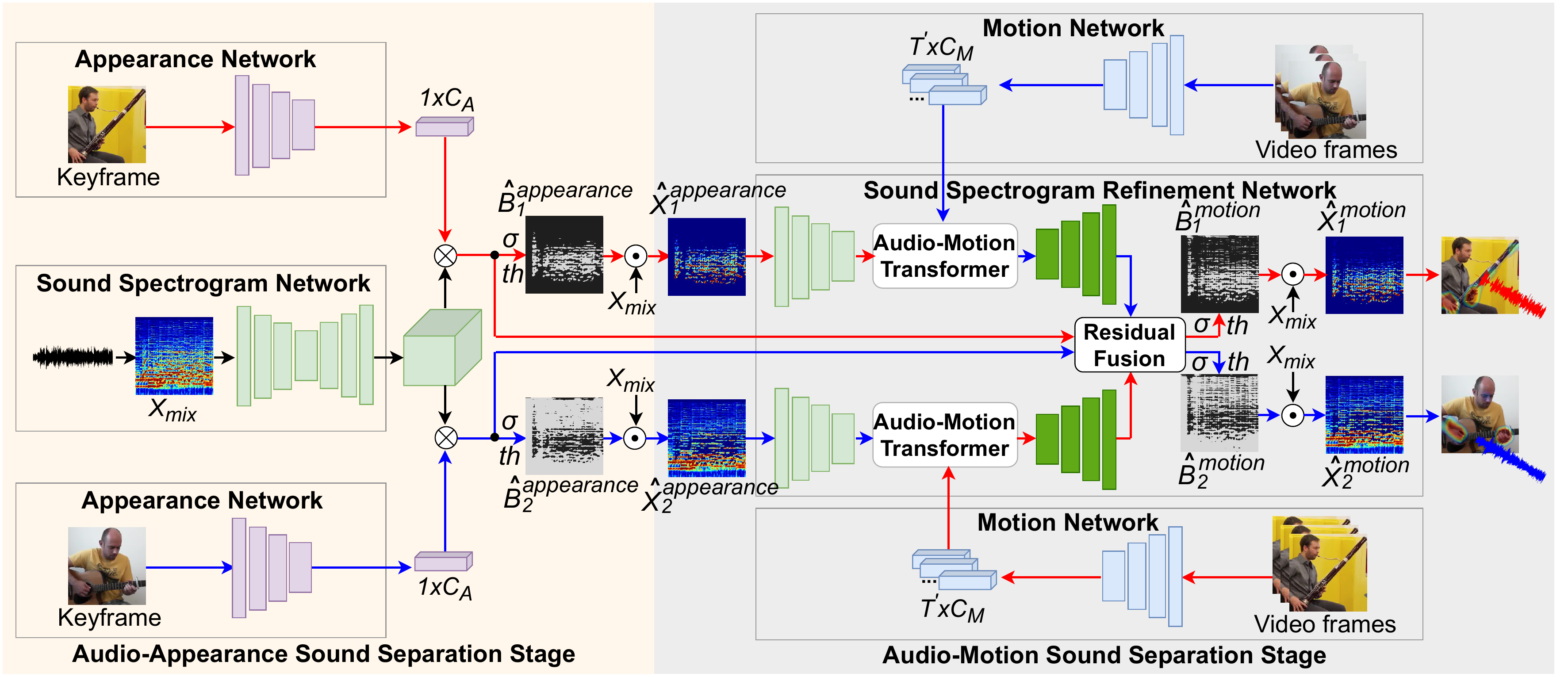}
   \caption{The overall architecture of the proposed Appearance and Motion network (AMnet). The Audio-Appearance stage encodes a video keyframe and a mixture audio spectrogram into an appearance feature vector and a spectrum feature volume, respectively. These are subsequently fused by calculating a weighted sum of spectrum features using appearance features as weights. The result is further converted to a binary mask and multiplied with the input spectrogram to produce the separated output. The Audio-Motion stage again encodes the spectrograms and the video sequence into feature representations. These are fused with the proposed Audio-Motion Transformer module, decoded, and passed to produce a refined mask, which is multiplied with the input mixture spectrogram to produce the final output.}
\label{fig:overview}
\end{figure*}

\section{Related Work}

\paragraph{\bf Audio-Visual Learning}

In recent years, leveraging constraints between different modalities, such as audio and vision, has been applied in various cross-model learning tasks. Aytar~\textit{et al.}~\cite{aytar2016soundnet} learned joint audio-visual embeddings to minimize the KL-divergence of their representations. Arandjelovic~\textit{et al.}~\cite{arandjelovic2017look,arandjelovic2018objects} associated the audio with vision by modeling their correspondence. More recently, researchers have demonstrated the works of audio-visual synchronization~\cite{owens2018audio,korbar2018cooperative,cheng2020look}, talking face generation~\cite{zhou2019talking}, audio-driven 3D facial animation~\cite{cudeiro2019capture}, audio-visual navigation~\cite{gan2020look}, and visual-to-auditory~\cite{hu2019listen,gan2020foley,zhou2020sep}.

\paragraph{\bf Visual Sound Separation}
Early works of sound source separation were mainly based on probabilistic methods~\cite{ghahramani1996factorial,roweis2001one,virtanen2007monaural,cichocki2009nonnegative}, while recent approaches utilise deep learning architectures \cite{simpson2015deep,hershey2016deep,chandna2017monoaural}. Despite of the substantial improvements, the general form of the problem is challenging and highly underdetermined. Visual sound separation is gaining increasing attentions recently. Ephrat~\textit{et al.}~\cite{ephrat2018looking} extracted face embeddings for speech separation. Similarly, Gao~\textit{et al.}~\cite{gao2019co,gao2018learning} applied object detection to facilitate source separation. Zhao~\textit{et al.}~\cite{zhao2018sound} proposed to separate sounds by a linear combination of semantic cues and sound spectrogram features. A subsequent work~\cite{zhao2019sound} introduced trajectory optical flows to the sound separation. Xu~\textit{et al.}~\cite{xu2019recursive} separated sounds by recursively removing large energy components from sound mixture. Gan~\textit{et al.}~\cite{gan2020music} associated body and finger movements with audio signals by learning a keypoint-based structured representation. While impressive, these methods either have limited capabilities to capture the motion cues or rely on prior knowledge (e.g. object detection, optical flows, or keypoints).

The works by Owens~\textit{et al.}~\cite{owens2018audio} and Zhu~\textit{et al.}~\cite{zhu2020visually} are most related to ours. \cite{owens2018audio} presented a classification-based audio-visual misalignment model to analyse multisensory features for sound separation. In~\cite{zhu2020visually}, the authors utilized visual features of all the sources to look for incorrectly assigned sound components between sources in a multi-stage manner. Our work learns new motion cues by mapping the audio and motions into a common embedding space, and separates sounds with the proposed AMnet, which specialises to appearance and the motion cues.

\paragraph{\bf Motion Representations of Video Sequence} 
Early works of video representations were largely based on handcrafted spatio-temporal features~\cite{laptev2005space,klaser2008spatio,wang2011action,wang2013action}. These have been recently shifted to deep neural networks, which can be roughly grouped into following categories: i) 2D CNN with summarized motions from dynamic images~\cite{bilen2016dynamic}; ii) 3D CNN on spatio-temporal video volume~\cite{tran2015learning}; iii) two-stream CNNs~\cite{simonyan2014two}, where motions are modeled from prior computed optical flows; iv) LSTM~\cite{donahue2015long}, Graph CNN~\cite{wang2018videos} and attention clusters~\cite{long2018attention} based techniques. These methods are proposed mainly for action recognition problem. In contrast, our goal is to model the visual motions of sound and further to facilitate the sound separation task.

\paragraph{\bf Sound Source Localization}
Visually identifying sound source location is another challenging task. Hershey~\textit{et al.}~\cite{hershey2000audio} utilized non-stationary Gaussian process to model audio-visual synchrony for locating sound sources. The subsequent work brought the ideas of canonical correlations~\cite{kidron2005pixels} and temporal coincidences~\cite{barzelay2007harmony}. More recent works, including semantics~\cite{arandjelovic2018objects,senocak2018learning,zhao2018sound,xu2019recursive,zhu2020separating}, trajectory optical flows~\cite{zhao2019sound}, misalignment~\cite{owens2018audio,korbar2018cooperative}, location masking~\cite{zhu2020visually}, spatial audios~\cite{yang2020telling}, and attention~\cite{arandjelovic2018objects,ramaswamy2020see,cheng2020look,zhu2020separating} based methods.

\section{Approach}

\subsection{Overview}
\label{sec:overview}

The input to the proposed system consists of a mixture audio and video sequences depicting the sound sources. The objective is to extract the component audio that corresponds to the sound source in the given video. Figure~\ref{fig:overview} illustrates the overall architecture of the proposed Appearance and Motion network (AMnet). The first part, called Audio-Appearance stage, performs source separation using pure appearance-based features (e.g.~object types). To this end, we first randomly extract a frame from the video sequence and encode it into a visual feature vector of dimension $\textit{C}_A$. The input mixture audio is converted to a $\textit{H}_S \times \textit{W}_S$ spectrogram image and then encoded into a feature volume of size $\textit{C}_S \times \textit{H}_S \times \textit{W}_S$. The obtained feature volume is multiplied channel-wise with the visual feature vector and converted to a $\textit{H}_S \times \textit{W}_S$ binary mask. The output of the Audio-Appearance stage is formed by multiplying the mixture spectrogram with the binary mask.

The obtained result is further passed to the Audio-Motion stage (see Figure \ref{fig:overview}), which refines the source separation using motion cues. The corresponding features are extracted by the Motion Network module, which encodes the video sequence into a motion representation of size $\textit{T}^{'}\times\textit{C}_M$, where $\textit{T}^{'}$ corresponds to the time dimension of the sequence. Here we do not make any pre-assumptions of the motion type (e.g.~human body motions). The subsequent Sound Spectrogram Refinement network (SSR) combines the motion features with the output spectrogram from the previous stage. The core part of the refinement network is the Audio-Motion Transformer (AMT) module that associates motion representations with the spectrogram features using a multimodal transformer architecture. The final output is formed by multiplying the original mixture spectrogram with a binary mask obtained from the refinement network. In addition, the Motion Network module provides an estimate of the sound source location in the video (see Figure~\ref{fig:vis_motion}). 

The Motion Network is trained using the new Audio-Motion Embedding (AME) framework. In AME, we learn mappings from both audio and video streams into a common embedding space (see Figure~\ref{fig:AVTA}), where distance would be correlated with temporal alignment of the input sequences. The following sections provide further details of the system parts and the training procedure. We start by describing the AME framework and related learning objectives. The detailed network architectures are provided in the supplementary material.

\subsection{Audio-Motion Embedding Framework}
\label{sec:AVTA}

\begin{figure}[tp]
    \centering
    \includegraphics[width=0.9\linewidth]{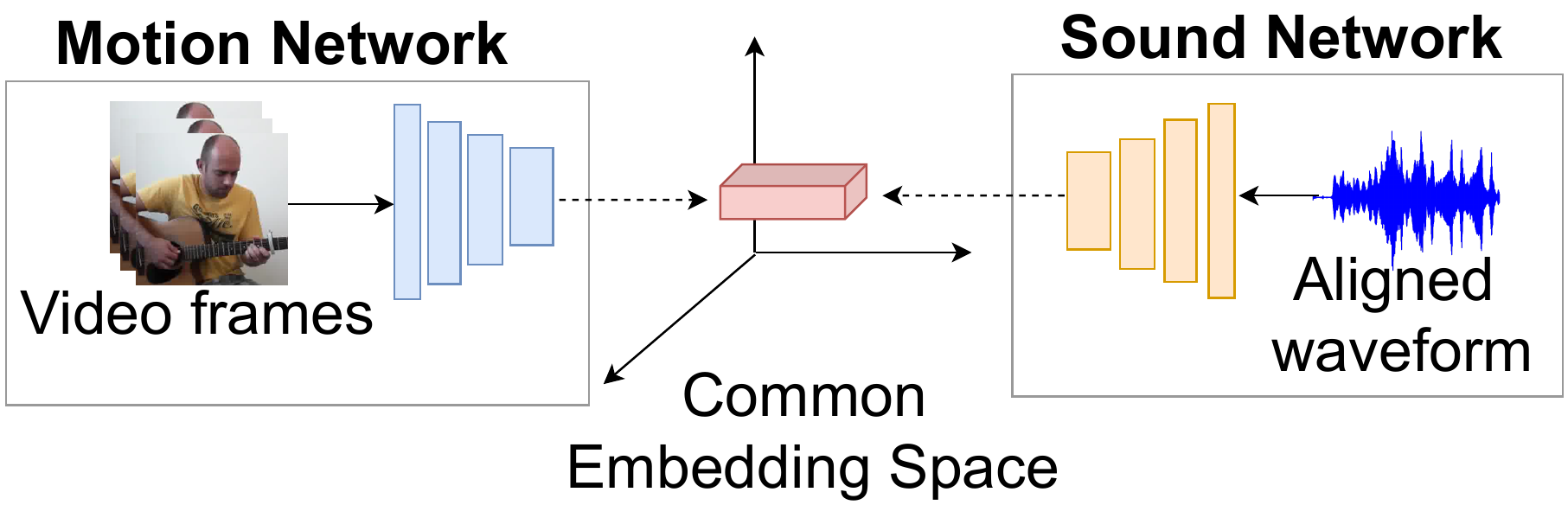}
   \caption{An illustration of the proposed Audio-Motion Embedding (AME) framework.}
\label{fig:AVTA}
\end{figure}

The proposed Audio-Motion Embedding (AME) framework exploits the natural correlation between the audio and motion of a natural video. The AME framework (see Figure~\ref{fig:AVTA}) consists of Motion Network and Sound Network, which map the motion and audio sequences into a common embedding space, respectively. We formulate the learning objectives to enforce small embedding distances between well synchronised streams and large distances for out-of-sync streams. We hypothesize that this learning objective encourages the Motion Network to focus mainly on the sound related motion features in the input video.

The AME maps the motions and audio into the common embedding space. The mappings are learned in the following manner: i) given a video clip $v$ with an aligned audio $x_{aligned}$, we generate a misaligned audio $x_{misaligned}$ by randomly shifting the waveform in time domain; ii) we encode the video stream, aligned audio, and the misaligned audio into embedding representations using the motion and sound networks, respectively; iii) we calculate the distances between the video embedding and the both audio embeddings, and formulate a cost function using a triplet loss approach~\cite{schroff2015facenet}; and iv) we optimise the embedding networks by minimizing the loss function over a large set of videos.

Previous works using audio-visual synchronization for self-supervised representation learning formulate the problem as a classification task where the system decides if the given audio and video are synchronized (1) or not (0) ~\cite{owens2018audio,korbar2018cooperative,cheng2020look}. Moreover, the information along the temporal dimension is mostly neglected by marginalizing the corresponding dimension with pooling operation. In contrast, we formulate the problem as a mapping from motion and audio domains to a common embedding space, where the distances correlate with the temporal alignment. Moreover, the representation retains the temporal dimension at the output features. Following paragraphs outline details of the embedding networks and the learning objective.

\paragraph{\bf Motion Network}
\label{sec:AVTAm}

The Motion Network $E_{M}$ maps the input video frames $v$ into a vector space. We forward the input video sequence $v$ of size $\textit{3}\times\textit{T}\times\textit{H}\times\textit{W}$ to a 3D version of Res18~\cite{he2016deep} and produce a representation $f_{M1}$ of size $\textit{C}_M\times{\textit{T}}^{'}\times\textit{H}^{'}\times\textit{W}^{'}$, where $T^{'}$=$T/4$, $H^{'}$=$H/16$, and $W^{'}$=$W/16$. With an additional 3D convolution, we obtain a single channel feature map $f_{M2}$ of size $\textit{1}\times{\textit{T}}^{'}\times\textit{H}^{'}\times\textit{W}^{'}$. We obtain the final embedding vector $f_{M3}$ of size $\textit{1}\times{\textit{T}}^{'}$ by applying a spatial average pooling.

\paragraph{\bf Sound Network}
The Sound Network $E_{S}$ maps the audio waveform into a common embedding space with the Motion Network $E_{M}$. Here we use Res18-1D architecture, which consists of a series of strided 1D convolutions, applied until the size of the output representation matches to the Motion Network output $f_{M3}$, i.e.~$\textit{1}\times{\textit{T}}^{'}$.

\paragraph{\bf Learning Objective}
We utilize the natural audio and motion temporal alignment to train the AME. Given a video clip, aligned audio, and misaligned audio \{$v$, $x_{aligned}$, $x_{misaligned}$\}, and their corresponding embeddings $E_{M}(v)$, $E_{S}(x_{aligned})$, and $E_{S}(x_{misaligned})$, we define a triplet loss function as follows:
\begin{equation}
    \begin{split}
        \mathcal{L}_{\textit{AME}} &= \max \Big( d \big( E_{M}(v), E_{S}(x_{aligned})\big) \\
            - \quad &d \big( E_{M}(v), E_{S}(x_{misaligned}) \big)
            + margin, 0 \Big)
    \end{split}
    \label{eq:triplet}    
\end{equation}
where $d$ measures the similarity between the motion and audio embeddings. We use $margin=2.0$ in all experiments. The embedding networks are optimized with respect to the loss function over a large set of input triplets.

\subsection{Audio-Appearance Sound Source Separation}
\label{sec:sep1}

The Audio-Appearance stage aims to perform source separation using object appearances. The stage consists of appearance network, sound spectrogram network, and sound source separation module. 

\paragraph{\bf Appearance Network}
The Appearance Network receives a random single frame from the input video and applies a dilated Res18-2D~\cite{he2016deep} to obtain a compact semantic representation. More specifically, given an input RGB image, the Appearance Network produces a representation $f_{A}$ of size $\textit{1}\times\textit{C}_{A}$ as the output of the last spatial average pooling layer.

\paragraph{\bf Sound Spectrogram Network}
The input audio waveform is first converted to a spectrogram presentation $X_{mix}$ using Short-time Fourier Transform (STFT). The Sound Spectrogram Network encodes $X_{mix}$ into a set of feature maps. The network is implemented using MobileNetV2 (MV2)~\cite{sandler2018mobilenetv2} architecture and it converts the input spectrogram of size $1\times\textit{H}_{S}\times\textit{W}_{S}$ to a feature map $f_{mix}$ of size $\textit{C}_{S}\times\textit{H}_{S}\times\textit{W}_{S}$. Note that the number of produced feature maps $\textit{C}_{S}$ is equal to the appearance feature vector dimension $\textit{C}_{A}$ in the previous section. 

\paragraph{\bf Sound Source Separation}

The sound source separation module utilises the feature maps $f_{A}$ and $f_{mix}$ to produce an estimate of the component audio corresponding to the input video. More specifically, 
\begin{equation}
    \begin{split}
        f^{appearance}_{S,n} = &f_{A,n} \otimes f_{mix}, \\
        \hat{B}^{appearance}_{n} = &th(\sigma (f^{appearance}_{S,n})), \\
        \hat{X}^{appearance}_{S,n} = &\hat{B}^{appearance}_{n} \odot X_{mix}
    \end{split}
    \label{eq:stage1}    
\end{equation}
where $\otimes$ and $\odot$ denote the channel-wise and element-wise product, respectively. $\sigma$ represents the sigmoid operation. $th(x)=1$ if $x>0.5$ and $0$ otherwise. $f_{A,n}$ is the appearance network output for the $n$-th source ($n$-th input video). The output spectrogram $\hat{X}^{appearance}_{S,n}$ is formulated by element-wise multiplying the binary mask $\hat{B}^{appearance}_{n}$ with the original mixture spectrogram $X_{mix}$.

\subsection{Audio-Motion Sound Source Separation}
\label{sec:sep2}
The Audio-Motion stage utilises the motion cues for the sound source separation. The stage contains four components: motion network (see Section~\ref{sec:AVTAm}), sound spectrogram refinement (SSR) network, audio-motion transformer (AMT), and residual fusion module. 

\paragraph{\bf Sound Spectrogram Refinement Network}
\label{sec:refinement}
The Sound Spectrogram Refinement (SSR) network is an encoder-decoder architecture, which consists of 7 down- and 7 up- convolutional layers followed by a BatchNorm layer and Leaky ReLU. The encoder ($SSR_{E}$) takes the spectrogram $\hat{X}^{appearance}_{S,n}$ from the Audio-Appearance stage and produces a feature representation $f^{motion,encoder}_{S,n}$ (the superscript \textit{motion} refers to the Audio-Motion stage). The encoder is followed by the Audio-Motion Transformer (AMT), which fuses $f^{motion,encoder}_{S,n}$ with the motion features $f_{M}$ ($f_{M1}$ following with a spatial pooling, which is produced by the Motion Network in AME presented in Section~\ref{sec:AVTA}). 

The transformed output is passed to the up-convolutional decoder ($SSR_{D}$) to produce residual spectrum representation of size $\textit{H}_{S}\times\textit{W}_{S}$. We adopt similar opponent filter approach as in \cite{zhu2020visually} to relocate the identified residual sound components from Audio-Appearance outputs to final corresponding outputs. This operation is performed by Residual Fusion Module (see below). More formally,
\begin{equation}
    \begin{split}
        f^{motion,encoder}_{S,n} = &SSR_{E}(\hat{X}^{appearance}_{S,n}), \\
        f^{AMT}_{S,n->m} = &AMT(f_{M,m}, f^{motion,encoder}_{S,n}), \\
        f^{motion,decoder}_{S,n->m} = &SSR_{D}(f^{AMT}_{S,n->m})
    \end{split}
    \label{eq:ssr}    
\end{equation}

\begin{figure}[tp]
    \centering
    \includegraphics[width=0.95\linewidth]{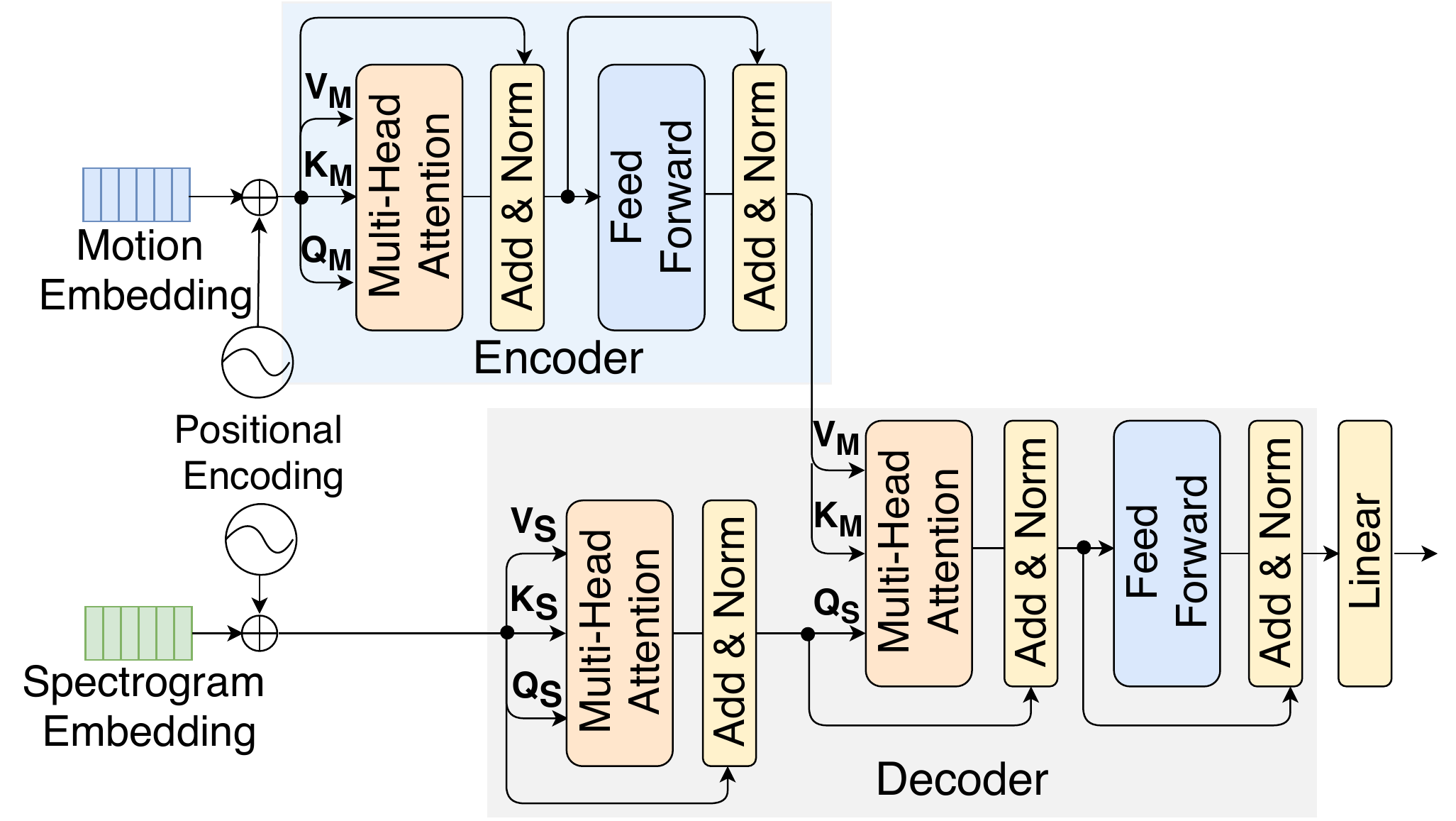}
   \caption{The architecture of the Audio-Motion Transformer.}
\label{fig:arch_transformer}
\end{figure}

\paragraph{\bf Audio-Motion Transformer}
\label{sec:avt}
The Audio-Motion Transformer (AMT) module is used to leverage the obtained motion cues (the motions of sound learned with AME) for sound source separation. The overall architecture follows~\cite{vaswani2017attention,iashin2020multi} and is visualised in Figure~\ref{fig:arch_transformer}. First the input embeddings are positional encoded~\cite{vaswani2017attention} to preserve ordering of the sequence. The following encoder and decoder modules are composed of stacked multi-head attention, point-wise, and fully connected layers.

The encoder applies $f_{M,m}$ of $m$-th source as query $Q_{M}$, key $K_{M}$, and value $V_{M}$ inputs. This type of multi-head attention which has identical $Q$, $K$, and $V$ is often referred as self-multi-head attention. Similarly, at the very beginning of the decoder, the AMT has a self-multi-head attention ($Q_{S}$, $K_{S}$, and $V_{S}$) of $n$-th sound features $f^{motion,encoder}_{S,n}$. After self-attention, the decoder infuses the motions into the sound components by an audio-motion multi-head attention, which considers the sound embedding $f^{AMT}_{S,n}$ of $n$-th source as the query $Q_{S}$, motions embeddings $f^{AMT}_{M,m}$ of $m$-th source as the key $K_{M}$ and value $V_{M}$, where the $m,n \in (1,\dots N), m \neq n$, and N is the number of sources.

\paragraph{\bf Residual Fusion Module}
\label{sec:residual}

The $SSR_{D}$ output $f^{motion,decoder}_{S,n->m}$ (Eq.~\ref{eq:ssr}) is interpreted as a residual spectrum. That is, the spectrum indicates the parts of the Audio-Appearance outputs (Eq.~\ref{eq:stage1}), which need to be reallocated. For instance $f^{motion,decoder}_{S,n->m}$ defines the component that should belong to source $m$ but is currently assigned to source $n$. Similar to \cite{zhu2020visually}, we relocate these components between the spectrograms, as follows
\begin{equation}
    \begin{split}
        f_{S,n} =&f^{appearance}_{S,n} \ominus f^{motion,decoder}_{S,n->m}, \\
        f_{S,m} =&f^{appearance}_{S,m} \oplus f^{motion,decoder}_{S,n->m}, \\
        &\hat{B}^{motion}_{n} = th(\sigma (f_{S,n})), \\
        &\hat{X}^{motion}_{S,n} = \hat{B}^{motion}_{n} \odot X_{mix}
    \end{split}
    \label{eq:residual}    
\end{equation}
where the $f^{appearance}_{S,n}$ is the output spectrum of Audio-Appearance stage for $n$-th source, $f^{motion,decoder}_{S,n->m}$ is the residual spectrum from sound $n$ to sound $m$. $\oplus$ and $\ominus$ denote the element-wise sum and subtraction, respectively. The obtained feature maps $f_{S,n}$ are passed through sigmoid and thresholding operations, and the final spectrogram output $\hat{X}^{motion}_{S,n}$ is formed by multiplying the original mixture spectrogram with the obtained result. An inverse STFT is applied to produce the final separated audio waveforms.

\subsection{Overall learning Objective}
\label{sec:objective}

We formulate the overall learning objective in terms of the binary masks $\hat{B}_{n}$, which are used to obtain the final output spectrograms (Eq.~\ref{eq:stage1} and~\ref{eq:residual}). The ground truth masks $B_{n}$ are formed as follows, 
\begin{equation}
    B_{n}(t, f)= [X_{n}(t, f) \geq X_{m}(t, f)]
\end{equation}
where $\forall m$ = $(1, \dots, N)$, (t, f) represents the time-frequency coordinates in the sound spectrogram $X$. The AMnet is trained by minimizing the binary cross entropy (BCE) loss between the estimated binary masks $\hat{B}_{n}$ and the ground-truth binary masks $B_{n}$,
\begin{equation}
    \mathcal{L} =\sum^N_{n=1} r_{1} \, \textit{BCE}(\hat{B}^{appearance}_{n}, B_{n}) + r_{2} \, \textit{BCE}(\hat{B}^{motion}_{n}, B_{n})
    \label{eq:objective}    
\end{equation}
where $\hat{B}^{appearance}_{n}$ and $\hat{B}^{motion}_{n}$ represent the predicted binary masks at Audio-Appearance and Audio-Motion stage, respectively. $r_{1}$ and $r_{2}$ are hyper-parameters which control the contribution of each loss factor.

\begin{figure}
    \centering
    \includegraphics[width=0.90\linewidth]{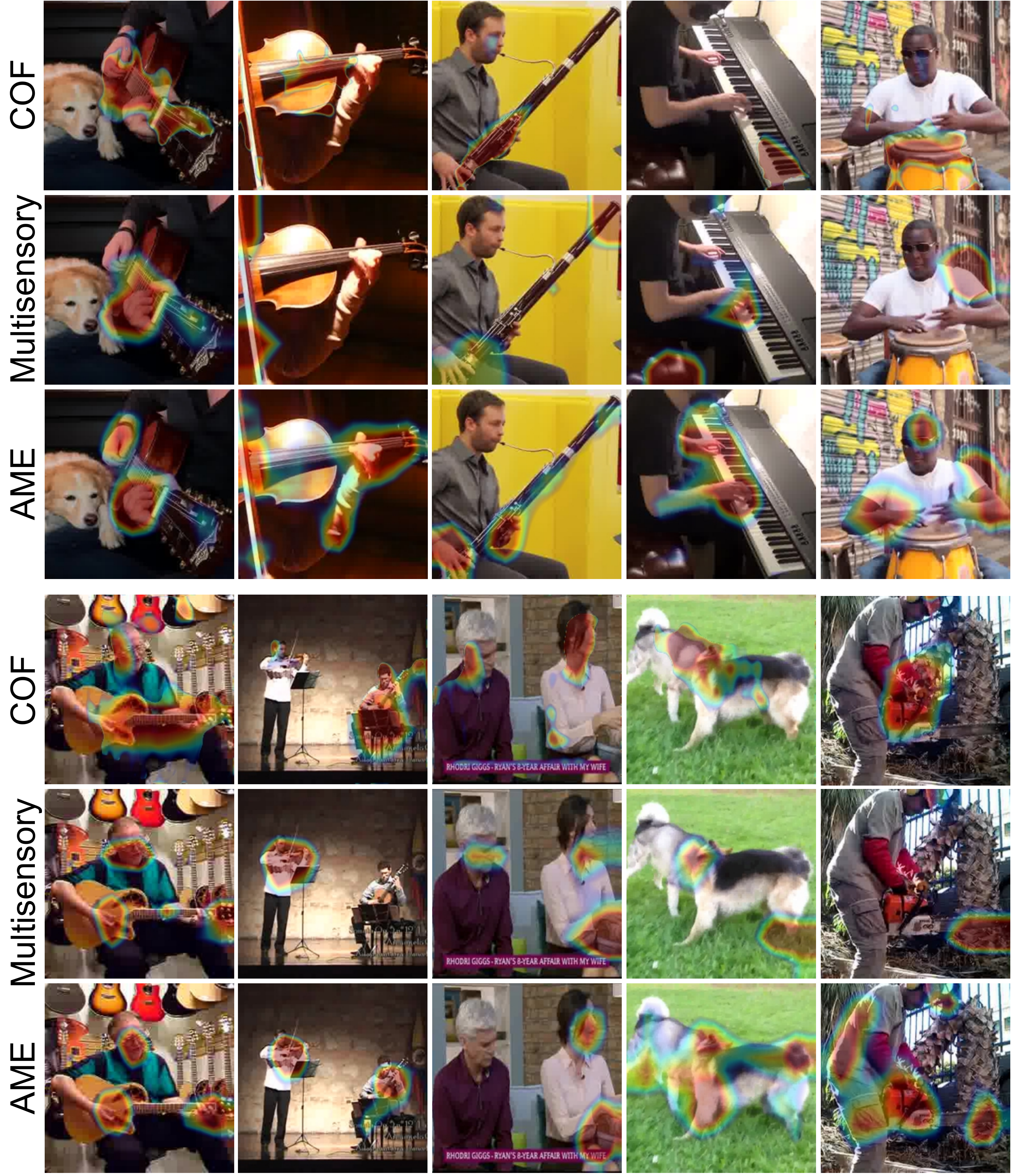}
   \caption{Visualization of the CAM responses for MUSIC-21 (top) and AVE (bottom) with our AME, COF~\cite{zhu2020visually}, and Multisensory~\cite{owens2018audio}.}
    \label{fig:vis_motion}
\end{figure}

\section{Experiments}

In this section, we start by assessing the Audio-Motion Embedding framework and continue with source separation experiments. The results are compared with the current state-of-the-art baseline methods. 

\subsection{Datasets and evaluation metrics}

We use MUSIC-21~\cite{zhao2019sound} and Audio-Visual Event (AVE) \cite{gemmeke2017audio,tian2018audio} datasets in our experiments. MUSIC-21~\cite{zhao2019sound} contains 1365 videos from 21 instrumental categories. We extract video frames at 8 fps and sub-sample audio streams at 11kHz. The AVE~\cite{tian2018audio} dataset, a subset of AudioSet~\cite{gemmeke2017audio}, contains 4143 10-second videos covering 28 event categories. The dataset covers a wide range of audio-visual events from different domains, e.g. human activities and vehicle sounds. For AVE, we use the full frame-rate (29.97 fps) and sub-sample audio signal at 22kHz. The sound separation performance is measured in terms of: Signal to Distortion Ratio (SDR), Signal to Interference Ratio (SIR), and Signal to Artifact Ratio (SAR). For measures, higher value indicates better performance.

\subsection{Implementation details}

The Audio-Motion Embedding (AME) framework is trained with MUSIC-21 dataset. We adopt random scaling, horizontal flipping, and cropping ($224\times224$) as the frame augmentation. A stream of $T=48$ frames is forwarded to the Motion Network $E_{M}$. We randomly crop 6-second audio clip and randomly shift the audio forward or backward by 1 to 7 seconds. In the source separation experiments, we follow the same setup as in \cite{gan2020music}. Both datasets are split into disjoint train, val (AVE), and test sets. The audio mixture is obtained by adding the audio tracks of N videos. For MUSIC-21, we use the same split as in \cite{gan2020music}. The input audio is converted to spectrogram using STFT with a hanning window of size 1022 and a hop lengths of 256 and 184 for MUSIC-21 and AVE datasets, respectively.

\setlength{\tabcolsep}{2pt}
\begin{table}
    \centering
    \begin{tabular}{lcc|cc|c}
        \hline\noalign{\smallskip}
        \multirow{2}{*}{Models $\backslash$ Metrics} &
            \multicolumn{2}{c}{MUSIC-21} &
            \multicolumn{2}{c}{AVE} &
            \multicolumn{1}{c}{UCF-101} \\
        \noalign{\smallskip}
         & cIoU & AUC & cIoU & AUC & Acc\\
        \hline
        Multisensory~\cite{owens2018audio} & 44.62 & 45.99 & 23.88 & 28.64 & 64.29\% \\
        AME (ours) & 67.18 & 54.58 & 25.62 & 29.81 & 71.69\% \\
        \hline
    \end{tabular}
    \caption{Quantitative localization results on MUSIC-21 and AVE datasets, and action recognition results on UCF-101 (split 1).}
    \label{tab:action_rec}
\end{table}
\setlength{\tabcolsep}{2pt}

\begin{table}
    \centering
    \begin{tabular}{lccc}
        \hline
        Models & SDR & SIR & SAR \\
        \hline
        Multisensory~\cite{owens2018audio} & 3.18 & 11.42 & 6.68 \\
        Sound of Pixels~\cite{zhao2018sound}$\star$  & 7.52 & 13.01 & 11.53 \\
        Co-Separation~\cite{gao2019co}$\star$ & 7.64 & 13.8 & 11.3 \\
        Sound of Motions~\cite{zhao2019sound}$\star$ & 8.31 & 14.82 &  13.11 \\
        Minus-Plus~\cite{xu2019recursive} & 9.15 & 15.38 & 12.11 \\
        Cascaded Opponent Filter~\cite{zhu2020visually} & 9.80 & 17.16 & 12.33 \\
        Music Gesture~\cite{gan2020music}$\star$ & 10.12 & 15.81 & - \\
        \hline
        AMnet (ours) & \bf 11.08 & \bf 18.00 & \bf 13.22 \\
        \hline
    \end{tabular}
    \caption{The source separation performance using mixtures of two sources from the MUSIC-21 dataset. The results indicated with $\star$ are obtained from \cite{gan2020music}.}
    \label{tab:sep_diff2}
\end{table}

\begin{table}
    \centering
    \begin{tabular}{lccc}
        \hline
        Models & SDR & SIR & SAR \\
        \hline
        Multisensory~\cite{owens2018audio} & 0.84 & 3.44 & 6.69 \\
        Sound of Pixels~\cite{zhao2018sound} & 1.21 & 7.08 & 6.84 \\
        Sound of Motions~\cite{zhao2019sound} & 1.48 & 7.41 & 7.39  \\
        Minus-Plus~\cite{xu2019recursive} & 1.96 & 7.95 & 8.08 \\
        Cascaded Opponent Filter~\cite{zhu2020visually} & 2.68 & 8.18 & 8.48 \\
        \hline
        AMnet (ours) &  \bf 3.71 & \bf 9.15 & \bf 11.00 \\
        \hline
    \end{tabular}
    \caption{The source separation performance for mixtures of two sources from the AVE dataset.}
    \label{tab:sep_av_syn}
\end{table}

\begin{table}
    \centering
    \begin{tabular}{lccc}
        \hline
        Models & SDR & SIR & SAR \\
        \hline
        Multisensory~\cite{owens2018audio} & -1.92 & 2.13 & 5.71 \\
        Sound of Pixels~\cite{zhao2018sound} & 2.31 & 9.34 & 5.77 \\
        Sound of Motions~\cite{zhao2019sound} & 2.77 & 10.20 & 5.81 \\
        Minus-Plus~\cite{xu2019recursive} & 3.36 & 9.22 & 7.15 \\
        Cascaded Opponent Filter~\cite{zhu2020visually} & 4.08 & 9.95 & 7.68 \\
        \hline
        AMnet (ours) & \bf 4.82 & \bf 11.75 & \bf 7.77 \\
        \hline
    \end{tabular}
    \caption{The source separation performance with mixtures of three sources from the MUSIC-21 dataset.}
    \label{tab:sep_diff3}
\end{table}

\subsection{Audio-Motion Embedding Framework}
\label{sec:exp_AVTA}

Although our main objective is to perform audio-visual source separation, we assess the AME based motion cues in three different motion related tasks: i) sound source localization; ii) action recognition; and iii) audio-visual sound source separation (in sec.~\ref{sec:exp_sep}). The motivation behind the additional comparisons is to provide wider picture of our model with respect to similar frameworks such as~\cite{owens2018audio}. The evaluation details are provided in supplementary material.

\paragraph{\bf Sound Source Localization}

For examining the ability of pinpointing the source location in the input stream, we measure the consensus Intersection over Union (cIoU) and Area Under Curve (AUC)~\cite{senocak2018learning} in Table~\ref{tab:action_rec} and visualize the source locations in Figure~\ref{fig:vis_motion}. Figure~\ref{fig:vis_motion} contains example results obtained using the Class Activation Map (CAM)~\cite{zhou2016learning}. More examples are provided in the supplementary material.

The results in Table~\ref{tab:action_rec} and Figure~\ref{fig:vis_motion} indicate that the sound related motion is nicely captured by the AME. For instance, the AME accurately captures both hands of the guitar player (first column in Figure~\ref{fig:vis_motion}). When playing a congas (fifth column in Figure~\ref{fig:vis_motion}), AME highlights hand and arm motions which are highly correlated with the output sound. The AVE dataset is more challenging due to the large scale of natural sounds it contains. Compared to the baseline models, our method can localize motions in various categories of videos (Figure~\ref{fig:vis_motion} bottom). For example, as shown in the fifth column (bottom Figure~\ref{fig:vis_motion}), our method detects both of the human body and the chainsaw motions as these motions occur always with the sounds. Moreover, as shown in the third column (bottom Figure~\ref{fig:vis_motion}), the woman on the right side is speaking while the man on the left is not. Even though they have similar appearance, the AME module localizes the motions precisely, whereas both COF~\cite{zhu2020visually} and Multisensory~\cite{owens2018audio} localize the head region of both the two persons. These results indicate that the AME mechanism facilitates the network to capture discriminative motions that correlated to the sound sources in the scenes instead of semantics.

\paragraph{\bf Action Recognition}

We further evaluate the performance of the AME motion cues in comparison with Multisensory~\cite{owens2018audio} for recognition tasks in Table~\ref{tab:action_rec} (last column). To study this, we
fine-tuned the methods (motion only) for action recognition using the UCF-101 dataset~\cite{soomro2012dataset}.

\begin{figure}
    \centering
    \includegraphics[width=1.0\linewidth]{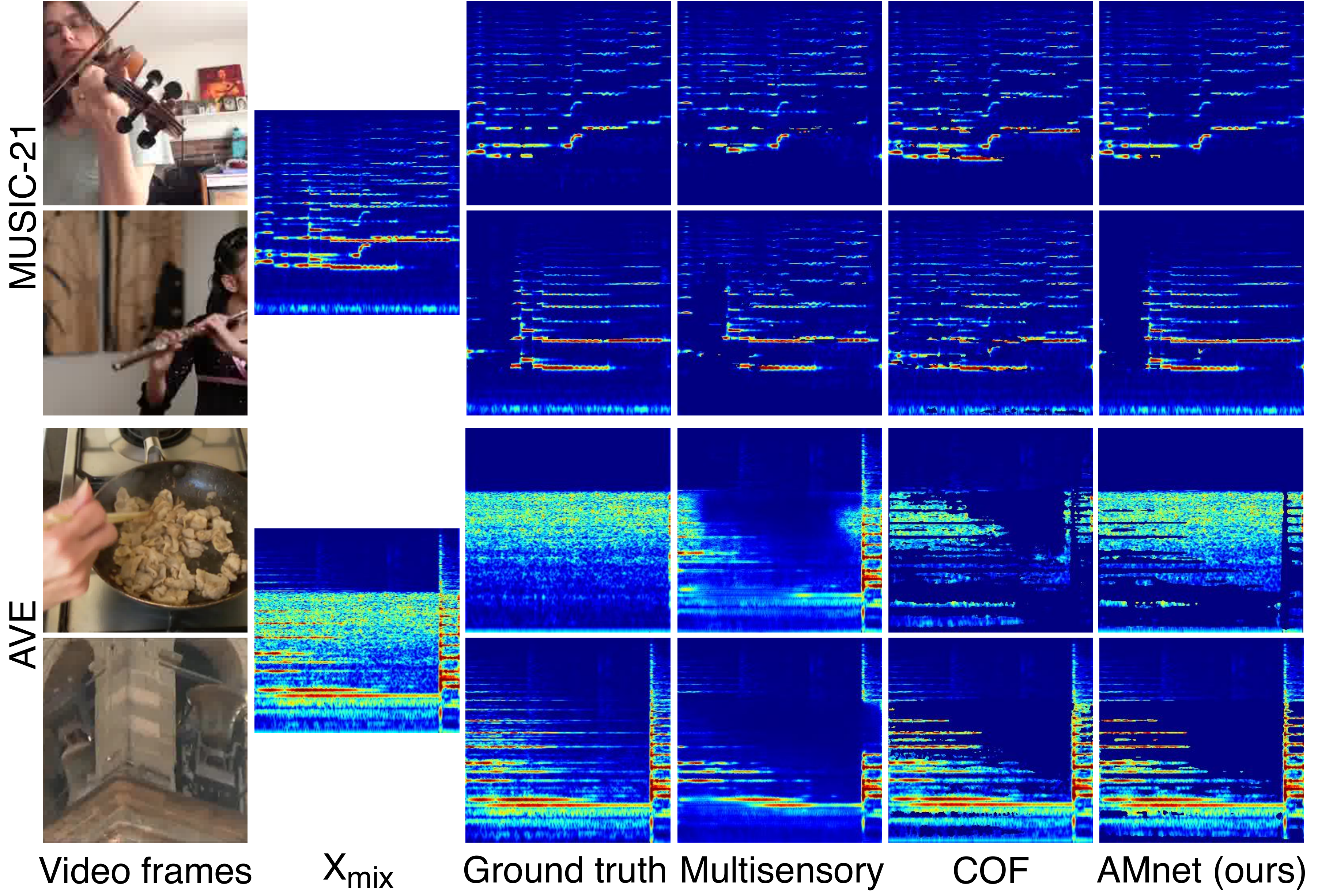}
   \caption{Visualization of the source separation results with the MUSIC-21 and AVE datasets.}
    \label{fig:vis_sep_diff}
\end{figure}

\subsection{Audio-Visual Sound Source Separation}
\label{sec:exp_sep}

\paragraph{\bf Separating Two Sound Sources}

Tables \ref{tab:sep_diff2} and \ref{tab:sep_av_syn} summarise the results for separating two sound sources using the proposed AMnet and recent baseline works \cite{owens2018audio,zhao2018sound,zhao2019sound,gao2019co,xu2019recursive,gan2020music,zhu2020visually}. The AMnet outperforms the baseline methods consistently and for most cases with a large margin. Compared to the closely related Multisensory~\cite{owens2018audio} and COF~\cite{zhu2020visually} works, we obtain 7.90dB and 1.28dB improvement on MUSIC-21 and 2.87dB and 1.03dB improvement on AVE, respectively. Figure \ref{fig:vis_sep_diff} contains qualitative examples, which clearly illustrate the differences. Additional examples are provided in the supplementary material. 

Interestingly, the proposed AMnet also outperforms the keypoint-assisted Music Gesture~\cite{gan2020music} model, which is particularly designed to exploit human body and finger dynamics. This result further indicates that the introduced AME model is able to capture fine-grained motion information from the sequences. Moreover, the transformer-based fusion with the motion features seems to provide a strong combination for the sound separation.

\paragraph{\bf Separating Three Sound Sources}

The separation task turns more difficult when the mixture contains more sources. To this end, we assess the methods by separating mixtures of three sources created from the MUSIC-21 dataset\footnote{Due to limited information, we were not able to exactly reproduce the results reported in \cite{zhao2019sound,gan2020music} for three sources and same instrument mixtures.}. Table~\ref{tab:sep_diff3} contains the results for AMnet and the baselines \cite{owens2018audio,zhao2018sound,zhao2019sound,xu2019recursive,zhu2020visually}. We can observe a clear drop in all performances compared to the two sources case (Table~\ref{tab:sep_diff2}). However, AMnet still outperforms the baselines with a clear margin.

\begin{figure}
    \centering
    \includegraphics[width=0.90\linewidth]{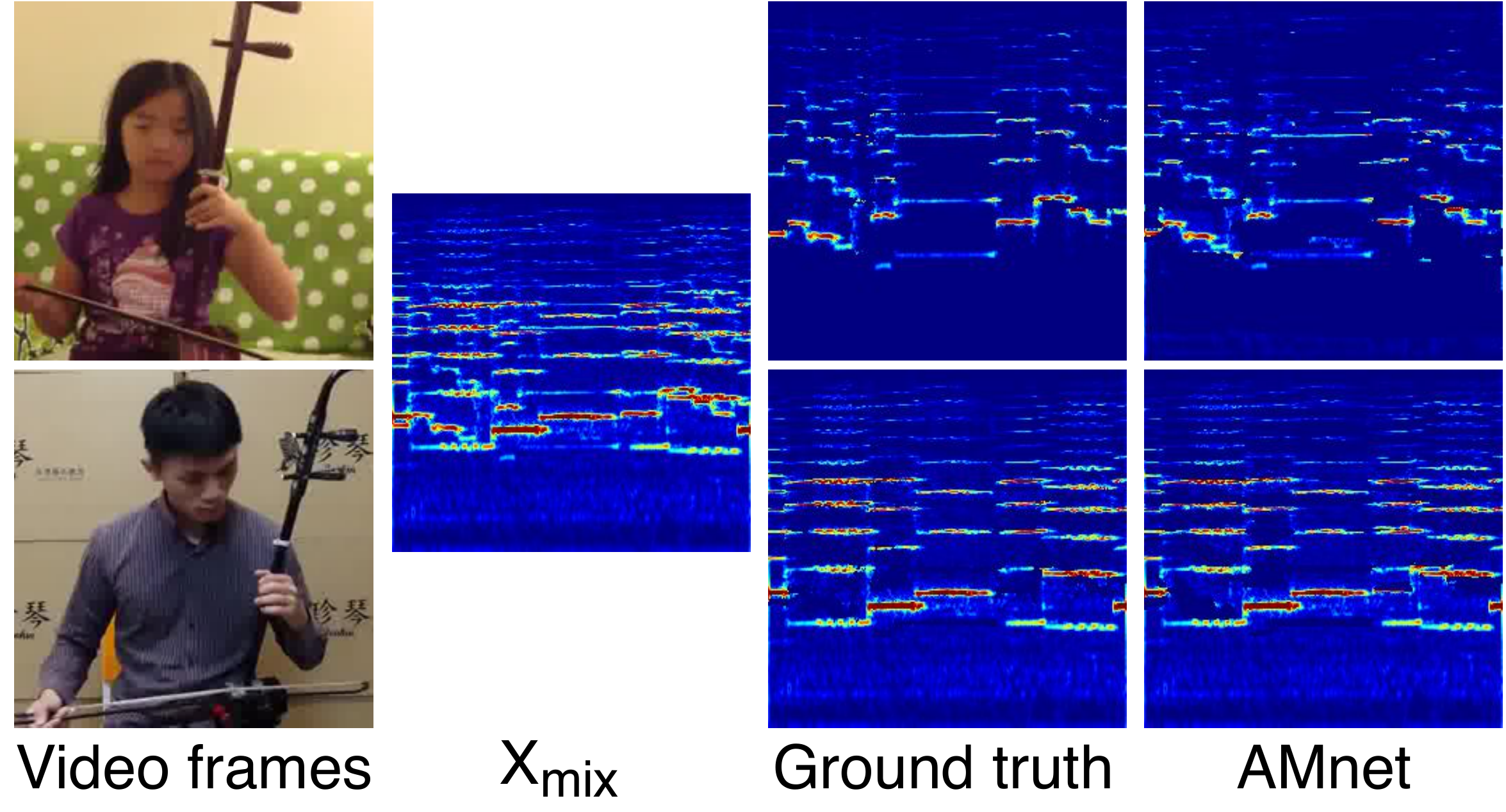}
   \caption{Illustration of the result for separating sound sources of the same type from MUSIC-21 dataset.}
    \label{fig:vis_sep_same}
\end{figure}

\setlength{\tabcolsep}{2pt}
\begin{table}
    \centering
    \begin{tabular}{lcccccc}
        \hline
        Models & 
        cello &
        erhu & 
        guzheng &
        pipa &
        xylophone \\
        \hline
        Copy-Paste & 0.55 & 0.12 & 0.29 & 0.24 & 0.30 \\
        SoP & -0.44 & -0.37 & -0.02 & -0.95 & 0.34 \\
        SoM & -4.04 & -0.92 & 0.62 & -2.34 & 0.25 \\
        Multisensory & -1.45 & 1.34 & -0.65 & -2.86 & -2.12 \\
        COF & -3.47 & 2.33 & -0.75 & 1.40 & 1.07 \\
        \hline
        AMnet (ours) & \bf 2.47 & \bf 3.91 & \bf 1.30 & \bf 2.28 & \bf 1.57 \\
        \hline
    \end{tabular}
    \caption{The source separation performance using mixtures of same instrument types from MUSIC-21 in terms of SDR.}
    \label{tab:sep_same}
\end{table}
\setlength{\tabcolsep}{2pt}

\paragraph{\bf Separating Sources of the Same Type}

In addition to the number of sources, the source type affects the performance. Particularly difficult case occurs when the sources are of the same type (e.g.~two cellos) and the appearance cues do not provide useful information. To this end, we pick video pairs with the same instrument type from MUSIC-21 and mix them for testing. We compare AMnet with Sound of Pixels (SoP)~\cite{zhao2018sound}, Sound of Motions (SoM)~\cite{zhao2019sound}, Multisensory~\cite{owens2018audio}, and COF~\cite{zhu2020visually}. The SoP is a pure appearance-based method, while the others utilise motion information in different ways. To gain further insight, we include a trivial ``Copy-Paste'' baseline that simply copies the input mixture spectrogram to the output. 

Table \ref{tab:sep_same} shows the results in terms of SDR for cello, erhu, guzheng, pipa, and xylophone categories. Interestingly, the Copy-Paste baseline outperforms comparison methods in most cases, whereas the proposed AMnet provides a clear improvement for all categories. The results indicate the challenges related to optical flow or dynamic image based motion modelling in self-supervised source separation. The classification-based multisensory features in~\cite{owens2018audio} suffer similar challenges. The motions of sound that learned through the audio-motion common embedding space seem to be the best option for capturing the motion cues in a self-supervised manner. 

\paragraph{\bf Ablation Study}

We conducted ablation study to investigate the performance of the Audio-Motion Transformer (AMT) and the role of the appearance and motions in our model. For this purpose, we repeat the source separation experiments with MUSIC-21 using different configurations. 

The AMT fuses the audio and motion for sound separation. To verify its efficacy, with a 1-stage model of using only motions to separate sounds, we replace the AMT module with the Early Fusion (EF) used in Multisensory~\cite{owens2018audio}. The comparison results are shown in Table~\ref{tab:avt}. Our proposed AMT module brings 1.11dB improvement in SDR.

Table~\ref{tab:stage} shows the results using only Motion, only Appearance, Appearance-Motion, and Appearance+Motion. For this case, the appearance results in better performance compared to using motion alone. Their combination in 1-stage improves the performance by 1.59dB in SDR. However, the full model AMnet provides clearly the best results suggesting that appearance and motion cues contain highly complementary information. Motion and appearance cues are different in nature and the AMnet lets each stage specialize in one type.

\begin{table}
    \centering
    \begin{tabular}{lcccc}
        \hline
        stages & Models & SDR & SIR & SAR \\
        \hline
        1 & Motion (EF) & 3.71 & 10.61 & 8.22 \\
        1 & Motion (AMT) & 4.82 & 12.04 & 8.46 \\
        \hline
    \end{tabular}
    \caption{The ablation results of using EF~\cite{owens2018audio} and AMT.}
    \label{tab:avt}
\end{table}

\begin{table}
    \centering
    \begin{tabular}{lcccc}
        \hline
        stages & Models & SDR & SIR & SAR \\
        \hline
        1 & Motion & 4.82 & 12.04 & 8.46 \\
        1 & Appearance & 7.40 & 12.88 & 11.00 \\
        1 & Appearance-Motion & 8.99 & 16.38 & 11.32 \\
        2 & Appearance+Motion (AMnet) & 11.08 & 18.00 & 13.22 \\
        \hline
    \end{tabular}
    \caption{The ablation results comparing the appearance and motion stages, and their combination. ``Appearance-Motion" represents a 1-stage model which fuses (by concatenation) the appearance and motions and continues to separate sound sources. ``Appearance+Motion" indicates a 2-stage model which includes appearance and motions at Audio-Appearance and Audio-Motion stage respectively.}
    \label{tab:stage}
\end{table}

\section{Conclusions}

In this paper, we show that the motion representations that learned through the Audio-Motion Embedding (AME) framework, together with the Appearance and Motion network (AMnet), obtain the new state-of-the-art results on visually guided sound source separation. The proposed AME approach results in better sound source localization and action recognition in comparison to baselines. The AMnet contains an Audio-Appearance and Audio-Motion stage, which specialise to appearance and motion cues, respectively. Our method, trained in a self-supervised manner, has no limitation on source types and outperforms the methods specifically designed to utilise human body motions. 

\paragraph{\bf Acknowledgement} This work is supported by the Academy of Finland (projects 327910 \& 324346).

{\small
\bibliographystyle{ieee_fullname}
\bibliography{ms}
}


\appendix
\section*{Supplementary Material}

The supplementary material is organized as follows: Section~\ref{sec:supp_vis} provides additional visualization of the source separation and localization; Section~\ref{sec:supp_net} contains additional details of the network architectures; and Section~\ref{sec:supp_opt} presents the optimization and evaluation configurations. The supplementary video illustrates the source separation comparison.

\section{Additional Qualitative Results}
This section provides additional qualitative results of the visual sound source separation and source localization results. The experimental setups are as explained in the main paper.  
\label{sec:supp_vis}

\begin{figure*}[!thp]
    \centering
    \includegraphics[width=1.0\linewidth]{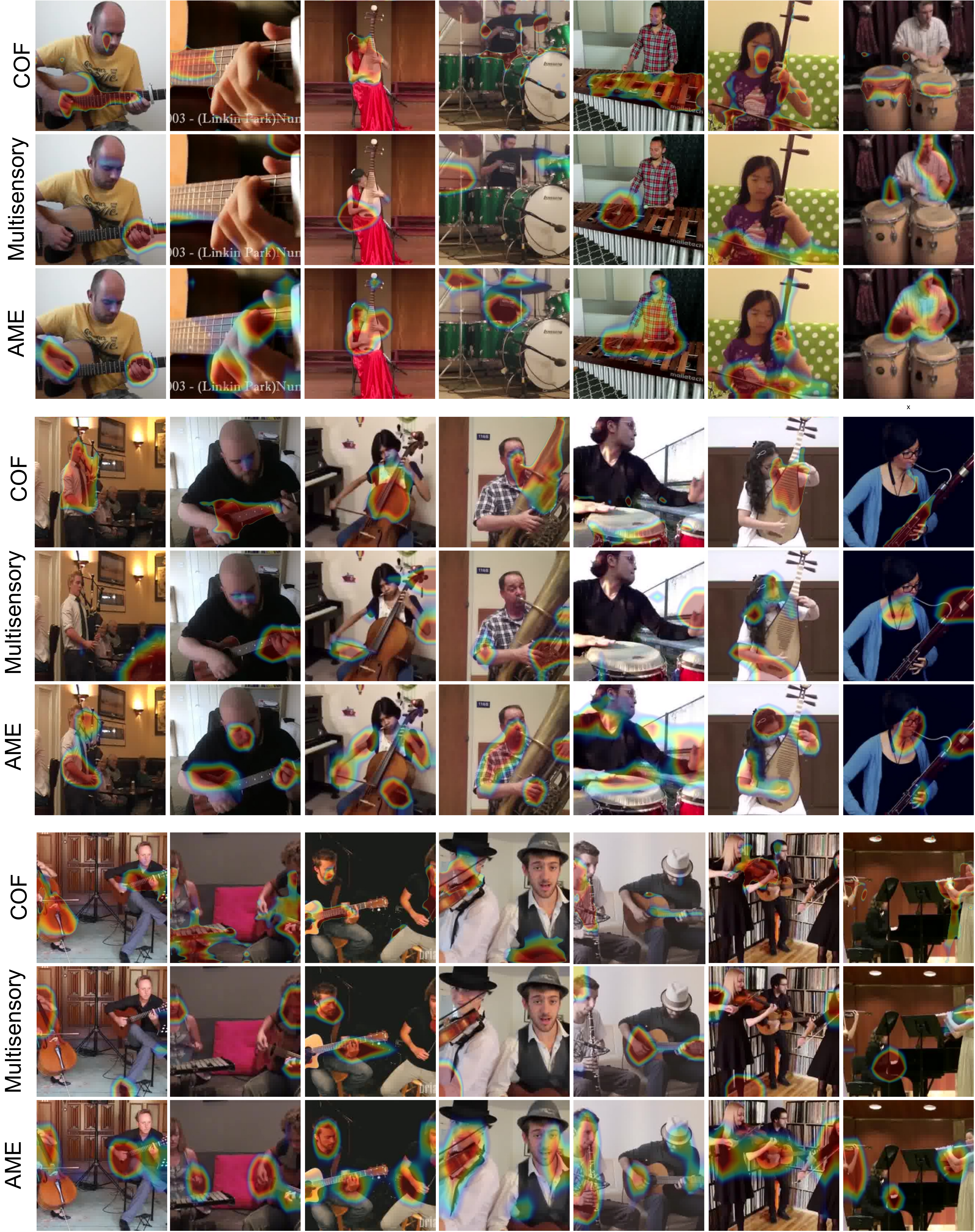}
   \caption{Visualization of the CAM responses with MUSIC-21 dataset for AME, COF, and Multisensory.}
\label{fig:vis_loc_MUSIC21}
\end{figure*}

\begin{figure*}[!thp]
    \centering
    \includegraphics[width=1.0\linewidth]{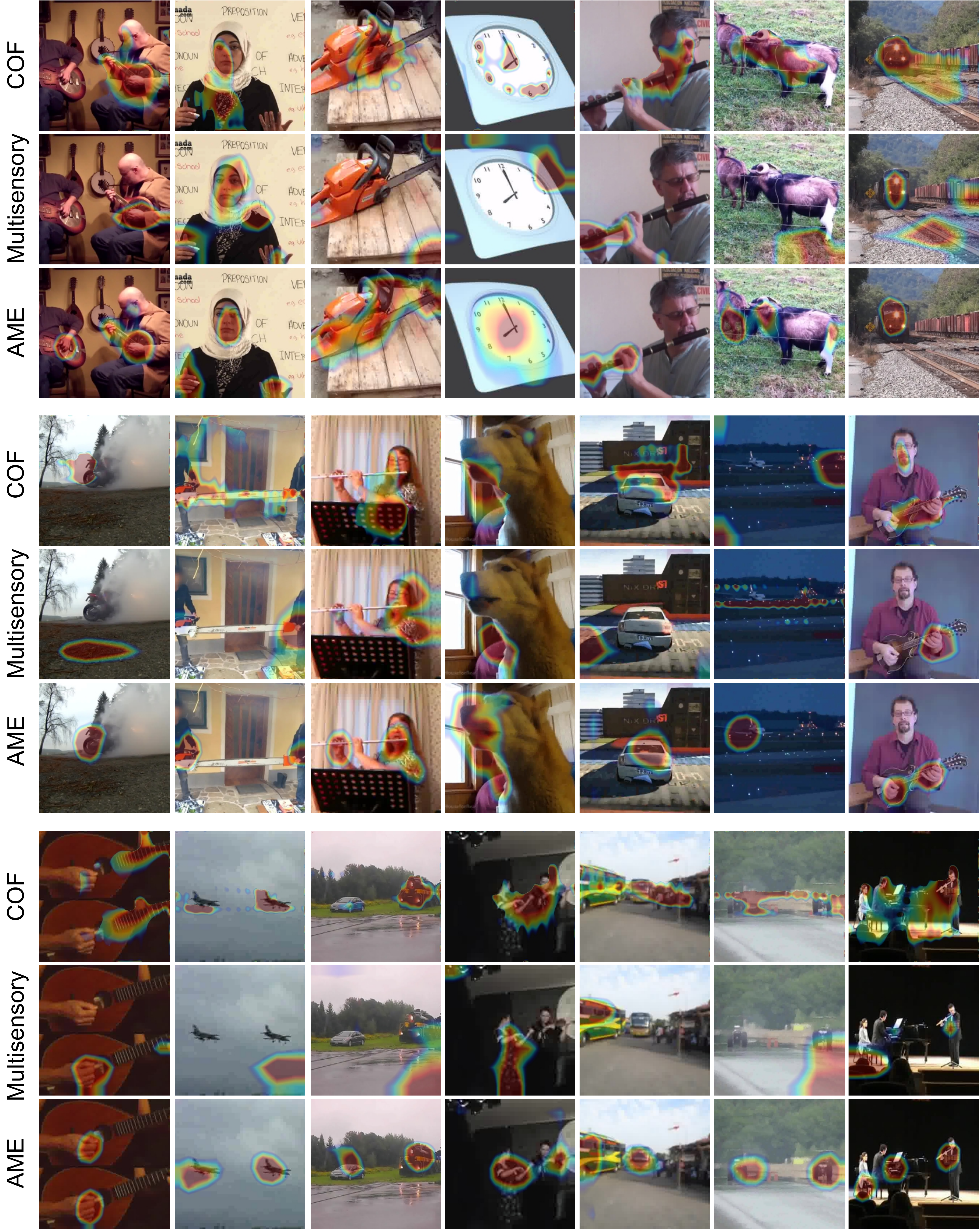}
   \caption{Visualization of the CAM responses with AVE dataset for AME, COF, and Multisensory.}
\label{fig:vis_loc_AVE}
\end{figure*}

\paragraph{Sound Source Localization}
Figures~\ref{fig:vis_loc_MUSIC21} and \ref{fig:vis_loc_AVE} provide additional qualitative results of the sound source localization with the proposed Audio-Motion Embedding (AME) framework, Cascaded Opponent Filter (COF)~\cite{zhu2020visually}, and Multisensory~\cite{owens2018audio} using MUSIC-21~\cite{zhao2019sound} and AVE~\cite{tian2018audio} datasets, respectively.

\begin{figure*}[!thp]
    \centering
    \includegraphics[width=1.0\linewidth]{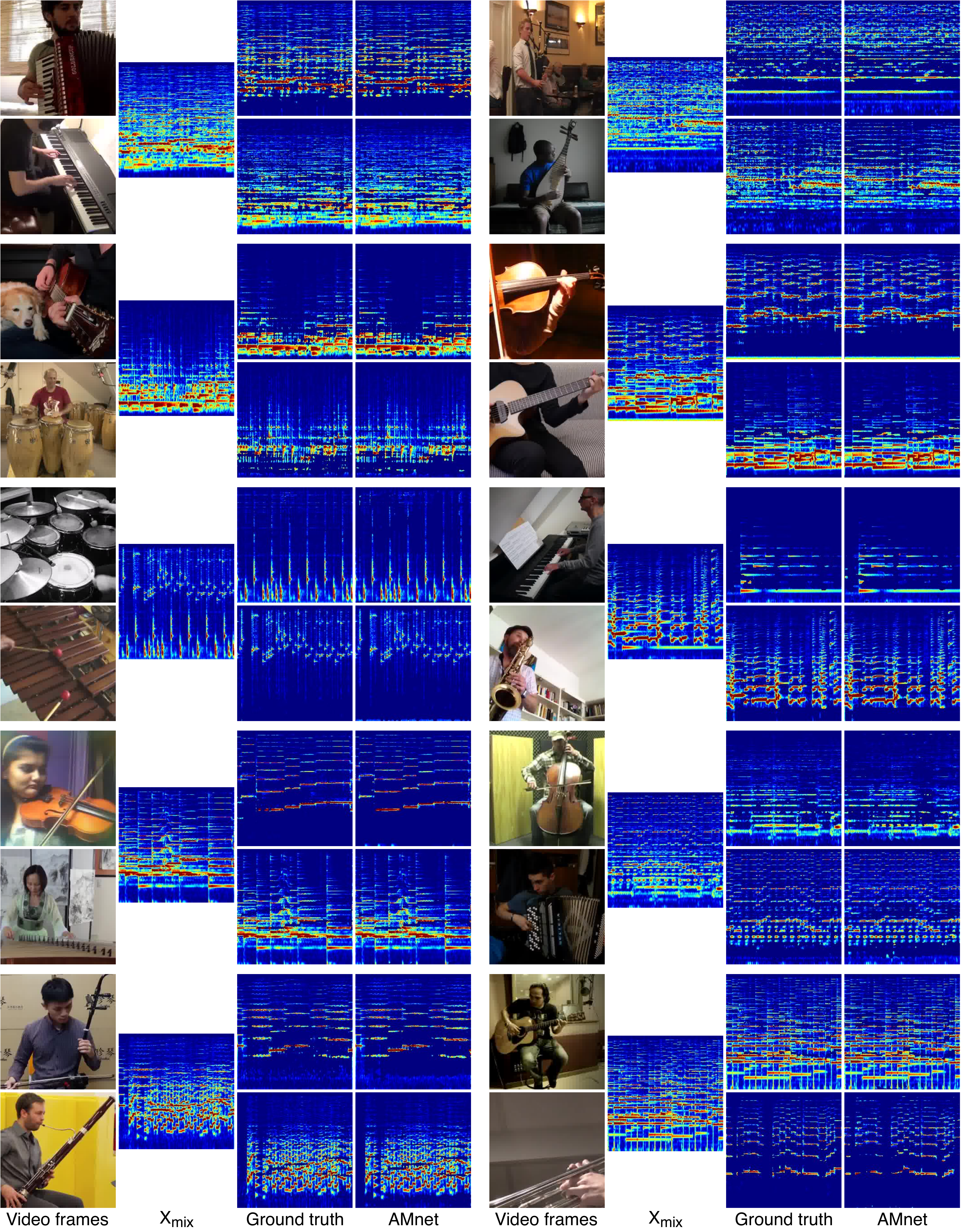}
   \caption{Visualization of the sound source separation results using AMnet with mixtures of two different sources from MUSIC-21 dataset.}
\label{fig:vis_sep_MUSIC21_2diff}
\end{figure*}

\begin{figure*}[!thp]
    \centering
    \includegraphics[width=1.0\linewidth]{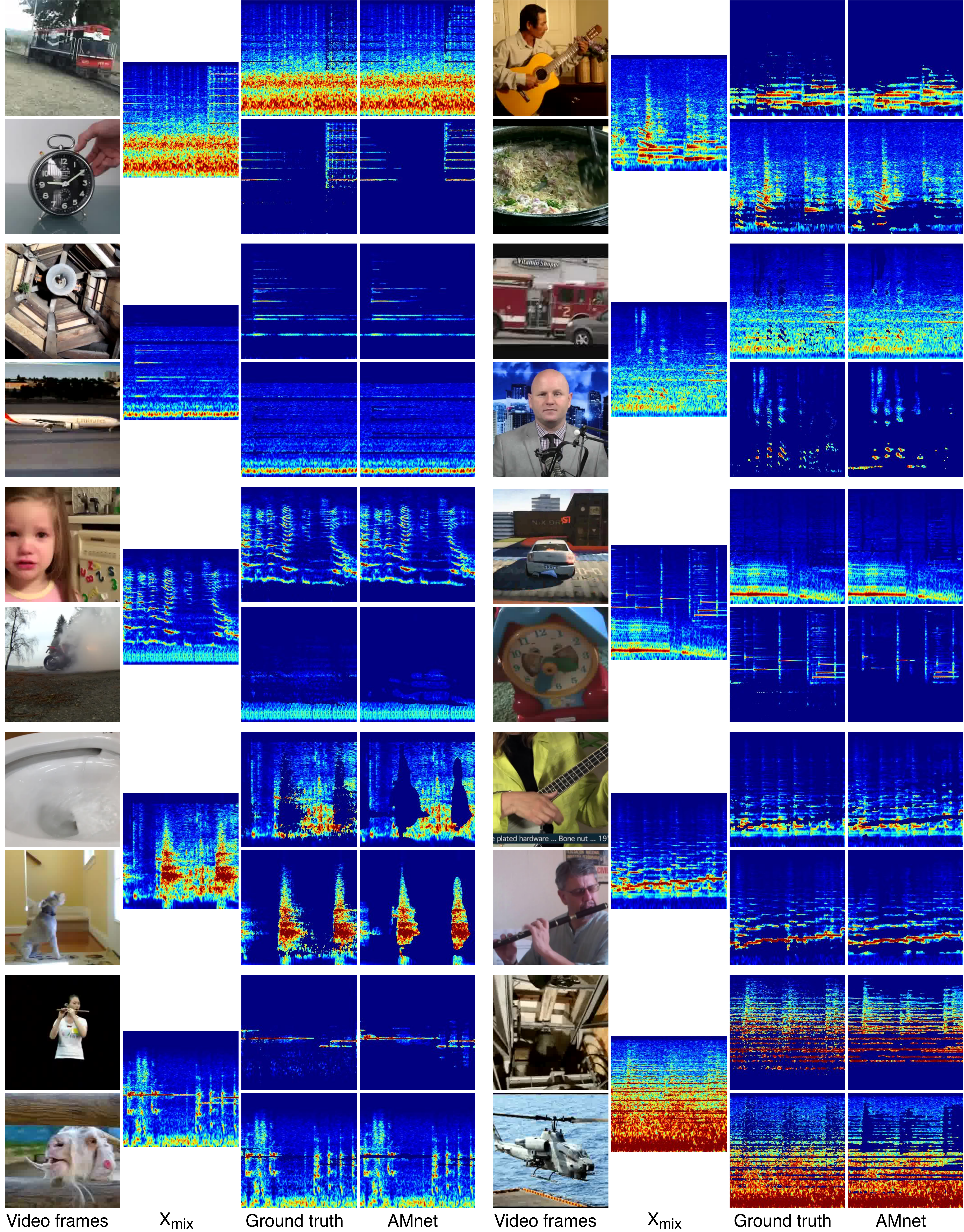}
   \caption{Visualization of the sound source separation results using AMnet with mixtures of two different sources from AVE dataset.}
\label{fig:vis_sep_AVE_2diff}
\end{figure*}

\begin{figure*}[!thp]
    \centering
    \includegraphics[width=1.0\linewidth]{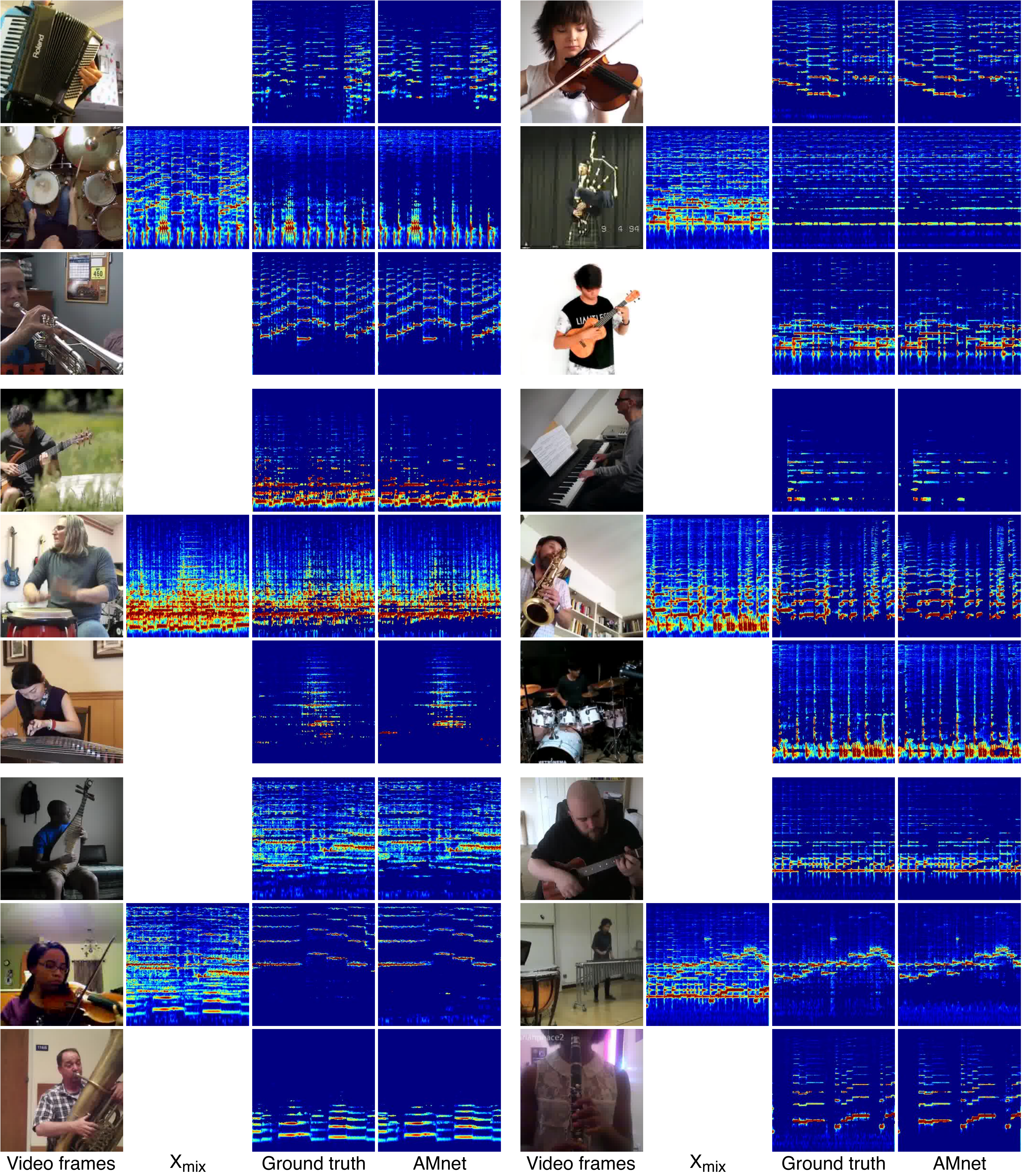}
   \caption{Visualization of the sound source separation results using AMnet with mixtures of three different sources from MUSIC-21 dataset.}
\label{fig:vis_sep_MUSIC21_3diff}
\end{figure*}

\begin{figure*}[!thp]
    \centering
    \includegraphics[width=1.0\linewidth]{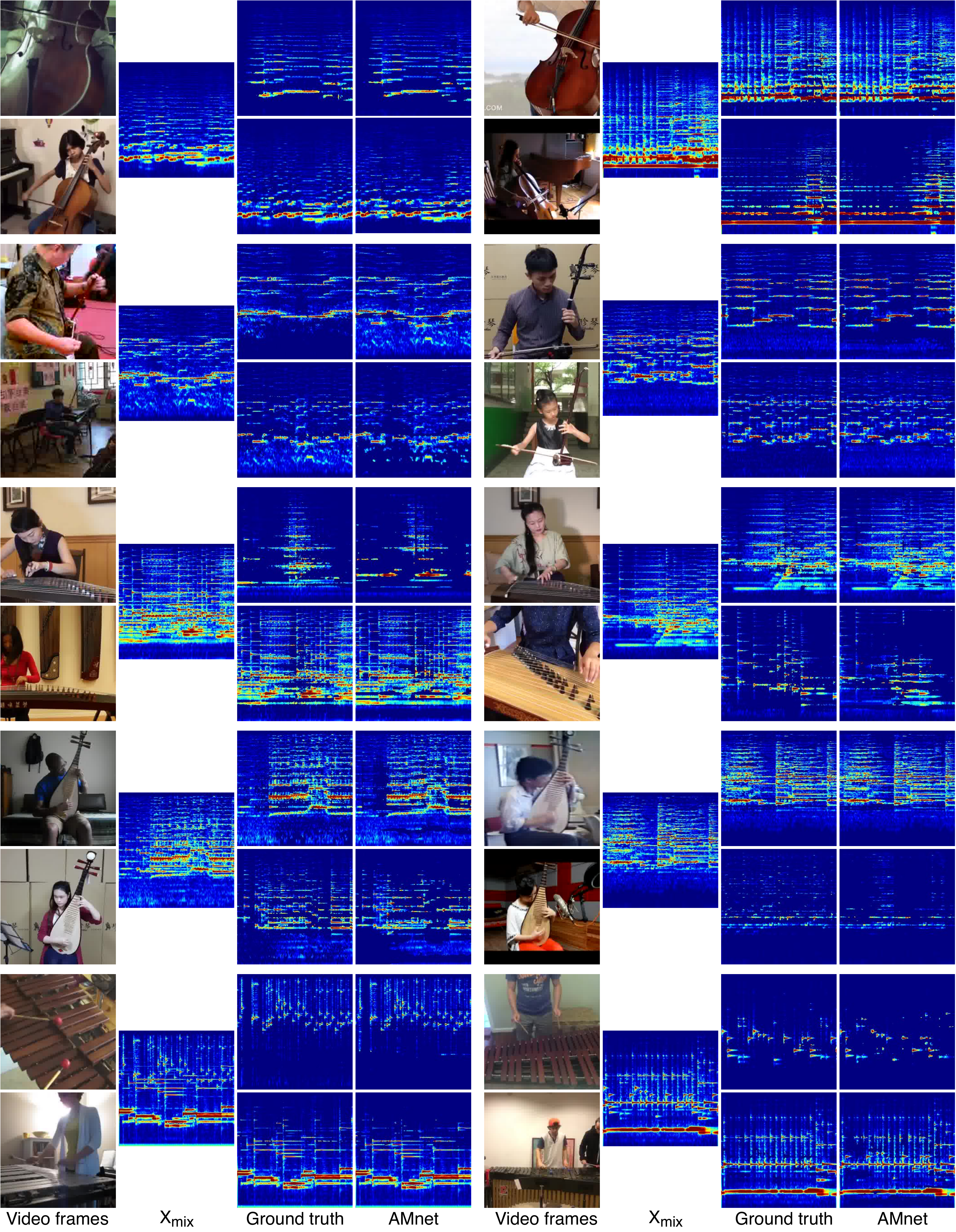}
   \caption{Visualization of the sound source separation results using AMnet with mixtures of two same type sources from MUSIC-21 dataset.}
\label{fig:vis_sep_MUSIC21_2same}
\end{figure*}

\paragraph{Visual Sound Source Separation}
Figures~\ref{fig:vis_sep_MUSIC21_2diff} and \ref{fig:vis_sep_AVE_2diff} present additional qualitative results of separating mixtures of two sound sources using AMnet from the MUSIC-21~\cite{zhao2019sound} and AVE~\cite{tian2018audio} datasets, respectively. Figure~\ref{fig:vis_sep_MUSIC21_3diff} shows results of separating mixtures of three sound sources from MUSIC-21. Figure~\ref{fig:vis_sep_MUSIC21_2same} contains results of separating sources of the same type from the MUSIC-21 dataset.

\section{Network Architectures}

\label{sec:supp_net}
This section provides additional details of the network structures and implementation.

\subsection{Audio-Motion Embedding Framework}
\label{sec:supp_AVTA}

\paragraph{\bf Motion Network}
\label{sec:supp_AVTA_Motion}
The Motion Network $E_{M}$ utilizes a 3D version of Res18 on the input video sequence of size $\textit{3}\times\textit{T}\times\textit{H}\times\textit{W}$, where $\textit{T}=\textit{48}$ and $\textit{H}=\textit{W}=224$. With the \textit{stride}=16 on spatial dimension and \textit{stride=4} on the temporal dimension, we yield the motion representation $f_{M1}$ of size $\textit{C}_M\times\textit{T}^{'}\times\textit{H}^{'}\times\textit{W}^{'}$, where $\textit{C}_{M}=\textit{512}$, $\textit{T}^{'}$=12 and $\textit{H}^{'}=\textit{W}^{'}=14$. With an additional 3D convolution, we obtain a single channel feature map $f_{M2}$ of size $\textit{1}\times\textit{T}^{'}\times\textit{H}^{'}\times\textit{W}^{'}$. Furthermore, we add a spatial average pooling over the $\textit{H}^{'}$ and $\textit{W}^{'}$ dimensions to achieve the final motion embedding vector $f_{M3}$ of size $\textit{1}\times\textit{T}^{'}$.

\paragraph{\bf Sound Network}
The Sound Network $E_{S}$ uses Res18-1D architecture to map the input audio waveform into a common vector space with the Motion Network. The Sound Network is composed of a series of 1D convolutions. A fractional poling and a 1D convolution layers are applied on top to obtain the final one channel embedding vector of size $\textit{1}\times\textit{T}^{'}$.

\subsection{Audio-Appearance Sound Source Separation}

\paragraph{\bf Appearance Network}

We adopt frame augmentation of random scaling, random horizontal flipping, and random cropping ($224\times224$) during training for all datasets. We apply a dilated Res18-2D with~\textit{dilation}=2 to obtain the appearance representations. For an input RGB image of size $\textit{3}\times\textit{H}\times\textit{W}$, we truncate the Res18-2D after \textit{stride}=16 and achieve the appearance feature of size $\textit{C}_{A}\times\textit{H}^{'}\times\textit{W}^{'}$, where $\textit{H}^{'}=\textit{W}^{'}=14$, $\textit{C}_{A}$ equals to $21$ and $28$ for MUSIC-21 and AVE datasets respectively. $\textit{C}_{A}$ represents the category numbers of dataset. By performing a spatial average pooling operation on the top, the Appearance Network produces the representation $f_{A}$ of size $\textit{1}\times\textit{C}_{A}$.

\paragraph{\bf Sound Spectrogram Network}
We firstly convert the input audio waveform into a spectrogram presentation $X_{mix}$ using Short-time Fourier Transform (STFT), and then forward the mixture spectrogram as the input of the Sound Spectrogram Network. The Sound Spectrogram Network is implemented using MobileNetV2 (MV2) architecture. The network converts the input spectrogram of size $1\times\textit{H}_{S}\times\textit{W}_{S}$ to a feature map $f_{mix}$ of size $\textit{C}_{S}\times\textit{H}_{S}\times\textit{W}_{S}$, where $\textit{H}_{S}=\textit{W}_{S}=256$, $\textit{C}_{S}$ equals to $21$ and $28$ for MUSIC-21 and AVE datasets respectively. Note that the number of produced feature maps $\textit{C}_{S}$ is equal to the appearance feature vector dimension $\textit{C}_{A}$ in the previous section.

\paragraph{\bf Sound Source Separation}
The sound source separation module combines the appearance representations $\textit{f}_{A,n}$ of $n$-th source with the sound spectrogram network output $f_{mix}$ using a linear combination to produce the spectrum features $f^{appearance}_{S,n}$ (the superscript \textit{appearance} refers to the Audio-Appearance stage) of size $\textit{1}\times\textit{256}\times\textit{256}$. With the \textit{sigmoid} and \textit{thresholding} ($th=\textit{0.5}$) operations, the spectrum features are converted to binary masks $\hat{B}^{appearance}_{n}$. The output spectrogram is formed by an element-wise multiplication between the binary mask and the original mixture spectrogram. We forward the output spectrograms $\hat{X}^{appeearance}_{S}$ of all the sources from the Audio-Appearance stage to the upcoming Audio-Motion stage as inputs.

\subsection{Audio-Motion Sound Source Separation}

\paragraph{\bf Motion Network}
The Motion Network in the Audio-Motion stage is pre-trained by the Audio-Motion Embedding (AME) framework in Section~\ref{sec:supp_AVTA_Motion}. We apply a spatial average pooling operation over the $\textit{H}^{'}$ and $\textit{W}^{'}$ dimension of the motion features $f_{M1}$ to obtain the motion representation of size $\textit{C}_{M}\times\textit{T}^{'}$, where $\textit{C}_{M}=\textit{512}$ and $\textit{T}^{'}=\textit{12}$.

\paragraph{\bf Sound Spectrogram Refinement Network}
The Sound Spectrogram Refinement (SSR) network takes the output spectrograms from the Audio-Appearance stage as inputs. The SSR has an encoder-decoder architecture. The encoder $\textit{SSR}_{E}$ processes the input spectrogram into sound features $f^{motion, encoder}_{S}$ (the superscript \textit{motion} refers to the Audio-Motion stage) of size $\textit{512}\times\textit{16}\times\textit{16}$. The encoder is followed by the Audio-Motion Transformer (AMT) module to fuse the motion and spectrogram features. We employ $8$ parallel heads attention layers in the AMT module. The following decoder $\textit{SSR}_{D}$ produces residual spectrum features $f^{motion, decoder}_{S,n->m}$ of size $\textit{1}\times\textit{256}\times\textit{256}$. We relocate the identified residual spectrum components from Audio-Appearance outputs to our final corresponding spectrum feature $f_{S}$ by using a Residual Fusion module (Eq.~4). With the \textit{sigmoid} and \textit{thresholding} ($th=\textit{0.5}$) operations, the spectrum features are converted to binary masks $\hat{B}^{motion}$. The output spectrogram $\hat{X}^{motion}_{S}$ is formulated by an element-wise multiplication between the resulted binary mask $\hat{B}^{motion}$ and the original mixture spectrogram $X_{mix}$. With an inverse STFT, we obtain the final separated audio waveforms.

\section{Implementation Details}
\label{sec:supp_opt}

\paragraph{\bf Optimization} 
The proposed method was implemented using Pytorch framework. We use stochastic gradient descent (SGD) with momentum 0.9, weight decay 1e-4, and batch size 10 for training AME and AMnet. The Appearance Network, pre-trained on ImageNet, uses a learning rate of 1e-4, while all other of modules are trained from scratch using a learning rate of 1e-3. 

\paragraph{\bf Evaluation} 

We assess the AME based motions cues in three different motion related tasks: i) sound source localization; ii) action recognition; and iii) audio-visual sound source separation. For all the evaluation metrics, higher value indicates better performance.

In order to give a quantitative evaluation of the AME motions, in addition to the qualitative visualizations, on the task of sound source localization, we measure the consensus Intersection over Union (cIoU) and Area Under Curve (AUC) metrics. Though with the fact that there is no direct dataset which has the ground truth of motion localization, we use the detected bounding boxes of mask r-cnn to indicate the coarse localization of sounding objects. 

For the action recognition task, we simply add a fully connected layer on top of the motion features for classifying the actions. We measure the performance by reporting the classification accuracy (Acc) on UCF-101~\cite{soomro2012dataset} dataset. 

The sound separation performance is measured in terms of: Signal to Distortion Ratio (SDR), Signal to Interference Ratio (SIR), and Signal to Artifact Ratio (SAR). SDR and SIR scores measure the separation accuracy. SAR captures only the absence of artifacts, hence can be high even if separation is poor.

\end{document}